Original Paper

# A Method to Learn Embedding of a Probabilistic Medical Knowledge Graph: Algorithm Development

Linfeng Li[1,2*], PhD; Peng Wang[3,4*], PhD; Yao Wang[2], MSc; Shenghui Wang[1], PhD; Jun Yan[2], PhD; Jinpeng Jiang[2], MSc; Buzhou Tang[5], PhD; Chengliang Wang[3], PhD; Yuting Liu[6], PhD

[1]Institute of Information Science, Beijing Jiaotong University, Beijing, China
[2]Yidu Cloud Technology Inc, Beijing, China
[3]College of Computer Science, Chongqing University, Chongqing, China
[4]Southwest Hospital, Chongqing, China
[5]Department of Computer Science, Harbin Institute of Technology Shenzhen Graduate School, Shenzhen, China
[6]School of Science, Beijing Jiaotong University, Beijing, China
[*]these authors contributed equally

**Corresponding Author:**
Yuting Liu, PhD
School of Science
Beijing Jiaotong University
No 3 Shangyuancun Haidian District
Beijing
China
Phone: 86 13810004230
Email: ytliu@bjtu.edu.cn

## Abstract

**Background:** Knowledge graph embedding is an effective semantic representation method for entities and relations in knowledge graphs. Several translation-based algorithms, including TransE, TransH, TransR, TransD, and TranSparse, have been proposed to learn effective embedding vectors from typical knowledge graphs in which the relations between head and tail entities are deterministic. However, in medical knowledge graphs, the relations between head and tail entities are inherently probabilistic. This difference introduces a challenge in embedding medical knowledge graphs.

**Objective:** We aimed to address the challenge of how to learn the probability values of triplets into representation vectors by making enhancements to existing TransX (where X is E, H, R, D, or Sparse) algorithms, including the following: (1) constructing a mapping function between the score value and the probability, and (2) introducing probability-based loss of triplets into the original margin-based loss function.

**Methods:** We performed the proposed PrTransX algorithm on a medical knowledge graph that we built from large-scale real-world electronic medical records data. We evaluated the embeddings using link prediction task.

**Results:** Compared with the corresponding TransX algorithms, the proposed PrTransX performed better than the TransX model in all evaluation indicators, achieving a higher proportion of corrected entities ranked in the top 10 and normalized discounted cumulative gain of the top 10 predicted tail entities, and lower mean rank.

**Conclusions:** The proposed PrTransX successfully incorporated the uncertainty of the knowledge triplets into the embedding vectors.

(*JMIR Med Inform 2020;8(5):e17645*)   doi: [10.2196/17645](10.2196/17645)

## Introduction

### Background

In medical fields, knowledge graphs (KGs) are the core underlying component of a clinical decision support system [1]. Clinical decision support system applications based on KGs have been reported in different scenarios, such as medicine recommendations [2] and drug-to-drug similarity measurements [3]. The KG is a graph-based knowledge representation method, which uses a set of (head, relation, tail) triplets to represent the various entities and their relationships in a domain. Each triplet is called as a fact as well. In KGs, nodes represent entities and edges represent relationships between entities. Medical KGs can be built either by human experts or by using unsupervised data mining from electronic medical records (EMRs). The first approach is too labor-intensive to be feasible for building large-scale KGs. Thus, unsupervised or semisupervised data mining from EMR data is a promising approach [4].

To learn effective knowledge representations, KG embedding has been proposed and gained massive attention, since the embedding vectors are easier to manipulate than the original symbolic entities and relations. The embedding algorithm maps the symbolic entities and relations into a continuous low-dimension vector space while preserving their semantic information. Different approaches to embed KGs by using translation-based learning embedding vectors are reported, such as TransE/H/R/D/Sparse [5-9] (noted as TransX hereafter). TransX algorithms learn embedding vectors from the deterministic facts in a KG. The learned embedding vectors help to improve performance of knowledge completion and other common natural language processing tasks [10]. In the medical field, embedding vectors are reported to be capable of improving diagnostic inference tasks [11].

### Related Approaches of Knowledge Graph Embedding

Bordes et al [5] implemented a translation-based algorithm (TransE) to model the (*h*,*r*,*t*) triplets in KGs. The score value of a given triplet is defined as the distance between h+r and t. The margin-based ranking criterion is defined as a loss function, and its target is to make the score value of a positive triplet be lower than that of a negative triplet by some margin. TransE features low model complexity while achieving relatively good predictive performance. Although the criticism has been made that TransE might learn similar vector representations for different tail entities in a 1-to-N relationship, the experiment results for TransH [6] and our study proved that such a flaw is not significant when the number of relations is small enough.

TransH [6] introduced relation-dependent hyperplanes to handle reflexive, 1-to-N, N-to-1, and N-to-N relations. The head and tail embedding vectors are mapped to the relation-dependent hyperplane, making it possible to project one entity into different projection vectors in different relations.

TransR [7] further extends the idea of the relation-specific projection by proposing to project an entity-embedding vector into a relation-specific vector space instead of a hyperplane. The introduction of the relation-specific space makes TransR more expressive at modeling differences among the relation and entities. Thus, TransR surpassed TransH in predicting the tail entities in many-to-many relations. CTransR is an extension of TransR that is designed to handle the differences in each relation. By clustering pairs within 1 relation, the implicit subtypes of a given relation are modeled as a cluster-specific relation vector. In the medical graph in this study, there is no such diversity in the relations.

TransD [8] replaces the relation-specific projection matrix with dynamic projection matrices for each entity-relation pair, thereby modeling various types and attributes among the entities. In addition, the dynamic projection matrices are constructed from projection vectors of head or tail entities and relations, requiring much fewer parameters and resulting in more efficient training. The proportion of corrected entities ranked in the top 10 (Hits@10) of predicting tails in many-to-many relations improved from 73.8% to 81.2% compared with CTransR.

Ji et al [9] argued that previous studies overfit simple relations and underfit complex relations, since relationships are heterogeneous and unbalanced. To resolve the challenge, they proposed the use of sparse matrices (TranSparse), either for each relation—that is, TranSparse(share)—or for each entity and relation—that is, TranSparse(separate). The sparse degree of each projection matrix is determined by the frequency of each relation or relation-entity pair in the training set. TranSparse(share) could be viewed as the counterpart of TransR, while TranSparse(separate) is the counterpart of TransD.

DIST_MULT [12] and ComplEx [13] do not use a distance-based score function; rather, they use a bilinear scoring function within a real vector space or a complex vector space. The loss function of DIST_MULT is the same as that of TransX, mentioned above, while the loss function of ComplEx is the negative log-likelihood of the logistic model. Experimental results show that the bilinear scoring function is more expressive, since DIST_MULT outperforms all TransXs at predicting tail entities in 1-to-N relations. ComplEx further surpassed DIST_MULT because the asymmetry in the products of complex embeddings helps to better express asymmetric relationships. However, because the score value is not distance based, it is difficult to map score values to probabilities.

He et al [14] proposed a density-based embedding method (KG2E) that embeds each entity and relation as a multidimensional Gaussian distribution. Such an embedding method is aimed at modeling the uncertainty of each entity and relation. However, KG2E did not achieve a better Hits@10 when predicting tails in many-to-many relations than CTransR [14]. We note that the uncertainty in KG2E is different from the probabilities of triplets use in this study. In KG2E, an entity is considered to have high certainty if it is contained in more triplets. In contrast, in this study we considered the probability as a metric of certainty of a triplet.

Fan et al [15] proposed a probabilistic belief embedding (PBE) model to measure the probability of each belief (*h*,*r*,*t*,*m*) in large-scale repositories. The notation *m* is the mention of the relation. The problem that PBE tried to solve is the most similar one to the problem we address in our study among all the related studies, which is embedding probability information from KGs into vectors. Apart from the extra element relation mention in

XSL•FO
RenderX

the quad, the key differences in the algorithms are as follows. First, in PBE, the probability of each triplet is calculated not only by using the embedding vectors of its head, relation, and tail, but also by using the embedding vectors of other triplets. In our study, the probability of each triplet depended only on the embedding vectors of itself. Second, in PBE, the softmax function is used to map distances into probabilities. The limitation of using the softmax is that it is difficult to model 1-to-many, many-to-1, and many-to-many relations. Consider that there are multiple tail entities, which are valid tail entities for a given head entity and relation. The ground truth probabilities of these triplets are 1. In Fan and colleagues' equations (8) and (14), the probability values of $Pr(r|h,r)$ are impossible to be trained to 1 for multiple valid tail entities, since the softmax function requires the sum of all probabilities to be 1.

Xiao et al [16] proposed a generative model (TransG) to learn multiple relation semantics. In medical fields, the relations do not contain different meanings as in a general knowledge base; thus, we did not necessarily consider TransG in this study.

Qian et al [17] argued that TransH/R/D/Sparse failed to learn that "various relations focus on different attributes of entities." Their model, TransAt, split the whole entity set into 3 according to the k-means: the capable candidate set of the head, the capable candidate set of the tail, and the rest. Furthermore, it set constraints on the distances of entities in intraset and interset pairs. In the medical KG that we built in our study, the candidate entities of the head and tail for a given relation are clearly defined by the relation itself. Thus, we did not consider TransAt in this study.

In contrast with the deterministic facts in general domain KGs, most facts in the medical domain are probabilistic. For example, considering the triplet (pneumonia, *disease_to_symptom*, fever), the symptom *fever* is common but not always present among patients with the disease *pneumonia* (code J18.901 in the *International Classification of Diseases, Tenth Revision* [ICD-10]). By counting the number of cooccurrences of *fever* and *pneumonia*, the conditional probability P(*symptom = fever|disease = pneumonia*) could be calculated from a real-world EMR data set.

Such a probabilistic nature is the unique challenge in embedding real-world EMR data-based medical KGs. In the KG on which TransX is designed, the label of each triplet is regarded as absolutely correct or wrong. During the training of TransX, a negative triplet (a triplet with an incorrect label) corresponding to each correct triplet is randomly sampled. Since the labels of triplets are binary in TransX, the training target is set as the score value of the positive triplet being lower than the score value of the negative triplet by some margin. In a medical KG, the label of a triplet—that is, the probability of the triplet—could indicate that (1) one triplet could be more likely (or unlikely) to appear than another or (2) the degree of certainty of each triplet is precisely expressed by the conditional probability. The probability of triplets is not considered by the TransX algorithms.

### Objective

We proposed PrTransX, based on the classical TransX algorithms, to learn embedding of a probabilistic KG. The main contributions of this study are as follows.

### Mapping Function Between Score Value and Probability

The score values of triplets in the existing TransX algorithms can reflect the geometric distance between the head and tail entities under a given relation. We proposed a function that can map the score value to the probability of a triplet and vice versa. This function both helps in translating the probability of a triplet to a target score during model training phase and enables users to convert a score value to a probability when predicating unforeseen links in model applications.

### Loss Function to Learn Embedding of a Probabilistic Knowledge Graph

Based on the mapping function, we introduced probability-based loss of triplets into the original margin-based loss function. Thus, the new loss function requires that the predicted probability of a triplet approximate its statistically probability on a training set. Finally, the triplet probabilistic information is learned into the embedding vectors.

### Evaluation of the Proposed Algorithms on Large-Scale Real-World Electronic Medical Records Data

We built probabilistic medical KGs from more than 10 million real-world EMR documents. KG embeddings were learned by the proposed algorithms.

## *Methods*

### Notation Overview

Table 1 explains the meanings of all mathematical symbols in this section.

### Building the Knowledge Graph From Real-World Electronic Medical Records

Real-world EMR data can be viewed as collections of visit records, each of which consists of all the medical records that are generated within 1 particular visit to a doctor by 1 patient, such as patient information, chief complaint, history of present illness, and medical orders. In each visit record, there are probably multiple medical entities. The term medical entity refers to a concrete instance of diagnosis, symptom, laboratory test, examination, medicine, and operation, such as the diagnosis *pneumonia, unspecified organism*, the symptom *cough*, the medicine *cefathiamidine*, and the examination *x-ray* in Figure 1.

**Table 1.** Notations used in the study.

| Symbols | Meaning |
| --- | --- |
| $h, r, t, h', r', t'$ | Head entities, relation, tail entities from positive triplet and negative triplet corresponding to positive triplet (marked as ') |
| $\Delta, \Delta'$ | Set of positive/negative triplets |
| h, r, t | Embedding vectors of head, relation, and tail entities |
| $h_p, t_p$ | Projection vectors of head and tail entities |
| $s_{(h,r,t)}$ | Score value of given triplet |
| $p_{(h,r,t)}$ | Probability of given triplet |
| $\Phi(), \Phi^{-1}$ | Mapping function between the score value and probability of triplets |
| $PL_{(h,r,t)}$ | Probability-based loss of given triplet |
| $\varepsilon_n$ | Probability value of negative triplet |
| $\varepsilon_p$ | Minimum probability value of positive triplet |
| $\lambda, K, \gamma$ | Scaling factors, margin parameters for loss function |
| $\alpha_r, \beta_r$ | Parameters for given relation $r$ |
| $[x]_+$ | The positive part of $x$ |
| $L_m$ | Margin-based loss function |
| $L$ | Loss function |

**Figure 1.** Workflow for extracting probabilistic knowledge triplets from real-world electronic medical record data. ICD10: *International Classification of Diseases, Tenth Revision*.

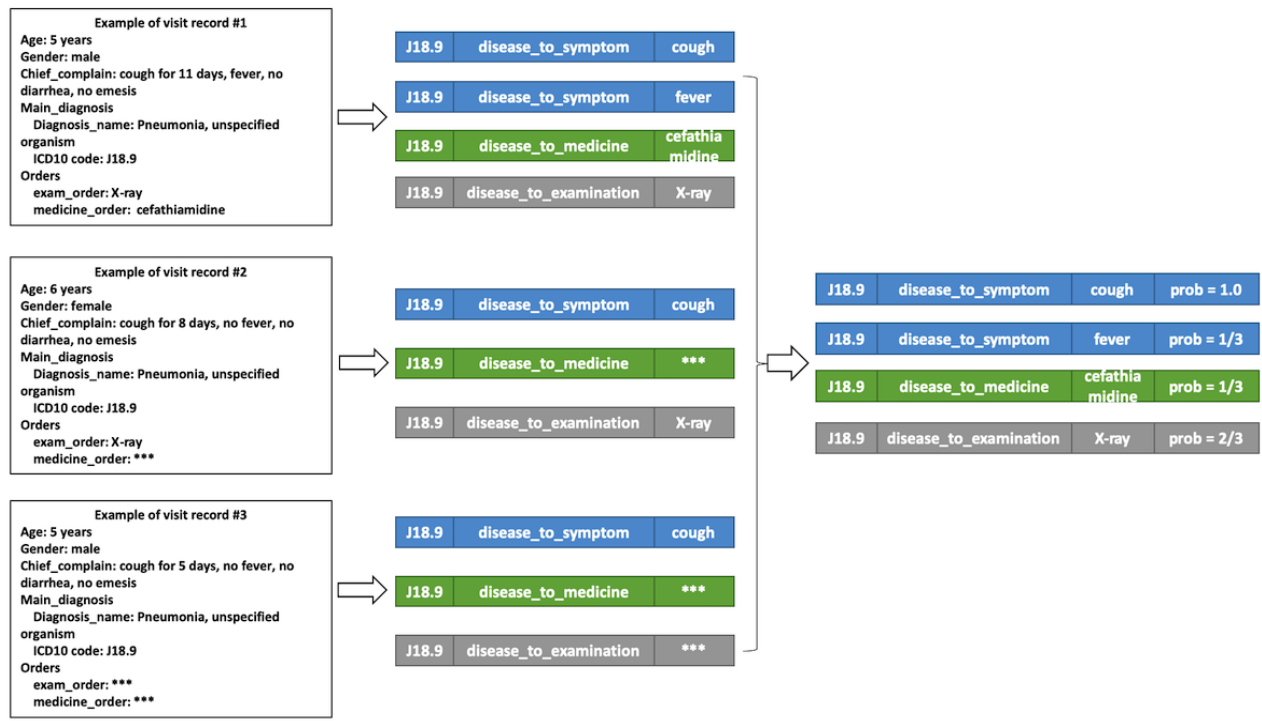

The relationships, which are expressed in (head entity, relation, tail entity) triplets, can be derived from the medical entities that occur in each single visit. By using the example visit above, several triplets can be derived, such as (J18.9, *disease_to_symptom*, cough) or (J18.9, *disease_to_medicine*, cefathiamidine).

The statistical probability of each triplet $(h,r,t)$ can be calculated by the equation $p_{(h,r,t)} = [N_{(h,r,t)}]/N_h$, where $N_{(h,r,t)}$ represents the number of visit records that would derive $(h,r,t)$ triplets, and $N_h$ represents the number of visit records that hold entity $h$. When $(h,r,t)$ is a valid triplet, the condition $p_{(h,r,t)}$ belongs to the set (0,1] holds; otherwise, $p_{(h,r,t)}=0$.

### Data Set Split and Ground Truth Setup

We used the *triplet group* as the minimum unit of separation between the training set and test set. One triplet group contained all triplets that shared the same head and relation. For example, all triplets whose head entity is disease C16.902 (gastric cancer not otherwise specified) and relation is *disease_to_symptom* constitute a triple group, containing (C16.902, *disease_to_symptom*, $s_1$), (C16.902, *disease_to_symptom*, $s_2$), ..., and (C16.902, *disease_to_symptom*, $s_N$), where $s_i$ indicates tail symptoms. One triplet group was split into either the training set or test set. In other words, for any given *h* and *r*, either all its tails were in training data or all were in test data.

In addition to the triplet group, we applied 2 rules during separation. First, for the relation *disease_to_XXX*, we randomly selected 20% of the triplet groups as the test set. Second, for the relation *upper_disease_to_lower_disease*, we included all triplets in the training set, since these relations are prior knowledge.

Noise is inevitable in real-world data. In medical cases, the most common noise in extracted triplets is due to unspecific triplets. Considering E11.901 (type 2 diabetes) as the head entity and *disease_to_laboratory* as the relation, triplets with routine laboratory test items (such as routine blood test) usually have higher probabilities than specific triplets (such as hemoglobin $A_{1c}$). However, according to medical knowledge, hemoglobin $A_{1c}$ is directly related to E11.901 (type 2 diabetes) as a diagnostic criterion, whereas routine blood test is not. Therefore, the original tail entities in a raw test set may be not really related to a head entity in medicine, and thus evaluating their performance cannot reveal the real ability to predict unknown medical knowledge.

To address the issue, medical experts manually label the ground truth tail entities according to clinical guidelines and the medical literature. During labeling, we also labeled the relevance level for the related tail entities, from 1 of the following values: strongly related (***), related (**), or weakly related (*). Since this is quite labor-intensive, we manually labeled only a randomly selected subset of the test data set. We named this subset the *evaluation data set*.

### Negative Triplets Sampling

In TransE [5], the negative triplets are created by randomly replacing the head or tail with a randomly selected entity. The authors of TransH [6] pointed out the false-negative issue and proposed that a Bernoulli distribution should be used to determine whether the head or tail should be replaced, which resulted in a lower number of false-negatives. We used sampling by using a Bernoulli distribution in the following TransH/R/D/Sparse algorithms.

In a medical KG that is built from real-world EMR data, background knowledge provides a more reliable way to sample negative triplets than the previous statistical method.

First, since the types of head and tail entities can be unambiguously determined by their relation, the valid candidate set for the replacement of the head or tail entity can be determined. For example, given the *disease_to_symptom* relation, the head entity must be the diagnosis type and the tail entity must be the symptom type.

Second, we can reduce the false-negatives by using medical knowledge. For example, the disease pneumonia (ICD-10: J18.901) is a hyponym of disease pneumonia unspecified (ICD-10: J18) in the ICD-10 [18] system. If the positive triplet is (pneumonia (ICD-10: J18.901), *disease_to_symptom*, symptom_x), we should not replace the head with any hyponym of pneumonia unspecified (ICD-10: J18) because diseases in the same ICD-10 class are likely to have similar properties.

In summary, by considering the entity type and domain knowledge, the number of false-negative triplets can be effectively reduced, against using the pure statistical method observed in previous work. We tested all the baseline models and proposed models using this enhanced negative sampling approach in our study.

### Mapping Function Between the Score Value and Probability

In previous TransX algorithms, the score value of a triplet was defined as the distance between $h_p+r$ and $t_p$, where $h_p$ and $t_p$ are vectors that are projected from the original embedding vectors h and t by some means, respectively. Such a score value calculation can be viewed as a function of (h,r,t), which is defined as $s_{(h,r,t)} = f_r(h,t) = \|h_p+r-t_p\|_{L1/L2}$.

Based on the above definition, the value range of the score value is [0, +inf). Furthermore, a value of 0 indicates a perfect match among (h,r,t), whereas a positive infinity value indicates that there is no relation between h and t under relation r. In the training of TransX, the target of the loss function was to simultaneously make the score value of the correct triplet of embeddings, which we noted as (h,r,t), as near to 0 as possible and to make the score value of the negative triplet, which we noted as (h′,r,t′), as large as possible.

In a medical KG, the label of a triplet is not absolutely correct or wrong, but rather it is a probability value that can be interpreted as the likelihood of the triplet being established. The probability value is 1 if a triplet holds under all conditions, and the probability value is 0 if the triplet does not hold under any valid condition.

It is clear that both the score value and probability can tell how likely a triplet is to be established. They are just different metrics. Therefore, we define a mapping function, Φ(), and its inverse function by Equation (1) in Figure 2.

**Figure 2.** Equations.

Equation (1)
$$p_{(h,r,t)} = \Phi(f_r(h,t)) = e^{-\lambda \cdot f_r(h,t)}$$
$$f_r(h,t) = \Phi^{-1}(p_{(h,r,t)}) = \frac{1}{\lambda} \ln \frac{1}{p_{(h,r,t)}}$$

where $\lambda > 0$ is a scaling coefficient

Equation (2)
$$p_{(h,r,t)} = \begin{cases} p_{(h,r,t)} \\ \varepsilon_p \end{cases} \quad \text{if } p_{(h,r,t)} < \varepsilon_p$$

where $0 < \varepsilon_n < \varepsilon_p$

Equation (3)
$$\mathcal{L}_m = \sum_{(h,r,t) \in \Delta} \sum_{(h',r,t') \in \Delta'} [s_{(h,r,t)} + \gamma - s_{(h',r,t')}]_+$$
$$= \sum_{(h,r,t) \in \Delta} \sum_{(h',r,t') \in \Delta'} [f_r(h,t) + \gamma - f_r(h',t')]_+$$

Equation (4)
$$PL_{(h,r,t)} = |\frac{1}{\lambda} \ln \frac{1}{p_{(h,r,t)}} - f_r(h,t)|$$

Equation (5)
$$PL_{(h',r,t')} = [\frac{1}{\lambda} \ln \frac{1}{\varepsilon_n} - f_r(h',t')]_+$$

Equation (6)
$$\mathcal{L} = \sum_{(h,r,t) \in \Delta} \sum_{(h',r,t') \in \Delta'} (e^{\frac{1}{K}[s_{(h,r,t)} + \gamma - s_{(h',r,t')}]_+}) \cdot (\alpha_r \cdot PL_{(h,r,t)} + \beta_r \cdot PL_{(h',r,t')})$$

where the hyperparameter K is a scaling factor to adjust the weight of the margin-based loss value; γ is the margin used to calculate the margin-based loss in the TransX model; $\alpha_r$ and $\beta_r$ are nonnegative coefficients used to adjust the weights between the positive probability-based loss and negative probability-based loss; the subscript r in $\alpha_r$ and $\beta_r$ means that these hyperparameters are set to different values for different relations.

To avoid dividing by 0, for a negative triplet t ($p_{(h',r,t')} = 0$), we defined an approximation as $p_{(h',r,t')} \approx \varepsilon_n, \varepsilon_n > 0$. To avoid the scenario that the probability of a valid triplet (h,r,t) is less than $\varepsilon_n$, we introduced the minimum probability of valid triplet as $\varepsilon_p$ in Equation (2), Figure 2).

The mathematical properties of Φ(*d*) facilitate learning effective embedding vectors from a medical KG. First, we 1-to-1 mapped the probability value in {x| belongs to the set [$\varepsilon_n$,1]} and the distance value in {*d*|*d* belongs to the set [0,ln(1/$\varepsilon_n$)]}. This made it possible to infer the quantitative probability of an unseen triplet only using the embedding vectors. Second, the calculation of the probability of (*h,r,t*) only depends on the embedding vectors (h,r,t) and not on any other embedding vectors. This means that the probability of one triplet could be independent of another. In 1-to-many and many-to-1 relations, it is common that multiple triplets are valid with high probabilities, and such independence avoids the shortcomings of using the softmax function to calculate the probability of a triplet in PBE [15].

### Loss Function to Learn Embedding of the Probabilistic Knowledge Graph

In previous TransX algorithms, the margin-based ranking criterion is defined as a loss function, which is expressed as Equation (3) in Figure 2. The training objective was to minimize the value of the loss function.

The major shortcoming of such a margin-based loss function is that it only requires $f_r(h,t)+\gamma \leq f_r(h',t')$, but it does not require that the predicted probability of each triplet approximate its statistical probability. To address the shortcoming, we defined the *probability-based loss of triplet*. Given a triplet (h,r,t) with probability $p_{(h,r,t)}$, we required that the targeted score value approximate the mapping of the actual probability $p_{(h,r,t)}$ by Φ(). Thus, we defined the loss as in Equation (4) in Figure 2. We defined the loss of the negative triplet by Equation (5) in Figure 2. We defined the loss function *L* of PrTransX as a combination of the margin-based loss, the probability-based loss of positive triplets, and the probability-based loss of negative triplets; see Equation (6) in Figure 2.

## Evaluation Protocol

To evaluate the performances of different algorithms, we used the link prediction task. The objective of link prediction in this study was to predict the tail entities using the heads and relations on the evaluation data set. For each triplet group in the evaluation data set, the embedding vectors that are trained by each algorithm can predict a sequence of tail entities in a valid entity type. An algorithm is considered to be better than another if it shows better results on evaluation metrics.

Referring to Bordes et al [5], we used Hits@10 and mean rank (the mean of those correctly predicted ranks) as evaluation metrics. The term *correctly predicted* or *hit* refers to the predicted item existing in the ground truth tail entities, regardless of the relevance level. Given that different tail entities would have different relevance levels with the head entities in medical knowledge, we also used the normalized discounted cumulative gain of the top 10 predicted tail entities (NDCG@10) [19] as a metric, since it measures whether an algorithm can rank more relevant tail entities in the front.

## Real-World Electronic Medical Records Data and Knowledge Graph

We performed this study using a dataset from the data platform and application platform of the Southwest Hospital in China. The platform is built based on a distributed computing architecture, and it is located on a private cloud in the hospital. The data platform and application platform aggregates medical data from EMRs, the hospital information system, laboratory information system, picture archiving and communication systems, and other isolated subsystems. The platform organizes all related medical data into visit-level and patient-level data. Visit-level data contain records from all subsystems for the same visit, such as the chief complaints and present illness histories from EMRs, examination orders and drug prescriptions from the hospital information system, and laboratory examination results from the laboratory information system. Patient-level data contain all visit data of the same patient.

## Results

### Data Set

We collected EMR records from 2015 to 2018 in the data set. The data set consisted of 3,767,198 patients and 16,217,270 visits. The entities from the data set were in 6 categories, described in Table 2. Among the entities, the disease entity was identified by its unique ICD-10 code [18]. ICD-10 hierarchical relationships are considered when extracting triplets. For example, assuming that the main diagnosis of a visit is C16.902 (gastric cancer not otherwise specified) and medicine m1 is prescribed by doctor, a list of triplets will be generated: (C16.902, *disease_to_medicine*, m1), (C16.9, *disease_to_medicine*, m1), and (C16, *disease_to_medicine*, m1).

**Table 2.** Description and distribution of relationships in the medical knowledge graph.

| relation_name | Source | Head entity type | Tail entity type | Triplet count |
| --- | --- | --- | --- | --- |
| *disease_to_medicine* | EMR[a] data set | Disease | Medicine | 74,835 |
| *disease_to_symptom* | EMR data set | Disease | Symptom | 53,885 |
| *disease_to_operation* | EMR data set | Disease | Operation | 13,292 |
| *disease_to_laboratory* | EMR data set | Disease | Laboratory | 71,805 |
| *disease_to_examination* | EMR data set | Disease | Examination | 38,061 |
| *upper_disease_to_lower_disease* | Domain knowledge: ICD-10[b] | Disease | Disease | 6455 |
| Total | —[c] | — | — | 258,333 |

[a]EMR: electronic medical record.
[b]ICD-10: International Classification of Diseases, Tenth Revision.
[c]Not applicable.

Based on the data set, we extracted triplets of 5 relationships. We added the relation *upper_disease_to_lower_disease* to the relation list as prior domain knowledge. Table 2 describes the features of the relationships. The original number of triplets in the data set was quite large because the probability of triplets was distributed as a long-tailed distribution. To reduce the noise from data and improve the training efficiency by reducing training data, we selected only the top 20 triplets (sorted in descending order of probability).

After separation, there were 205,877 triplets from 21,327 triplet groups in the training set, and 49,756 triplets from 4547 triplet groups in the test set.

There were in total 335 triplets from 25 triplet groups in the evaluation data set. We distributed the 25 triplet groups evenly among the 5 relationships to be evaluated.

### Baselines and Implementation

All translational distance-based TransX models could be extended to be PrTransX. Thus, the evaluation target was to compare the performances of the TransX methods and their corresponding PrTransX methods. We selected some of the translational distance-based TransX models: TransE [5], TransH [6], TransR [7], TransD [8], and TranSparse [9].

In addition to TransX, we also examined whether probabilistic inference algorithms, such as the naive Bayes, are feasible in this application. The conclusion was that they were not. To predict tail entities—that is, to calculate $P(t = t_i | h = h_0, r = r_0)$ for each candidate tail entity—the conditional likelihood $P(h = h_0 | t = t_i, r = r_0)$ must be available from the training set. However, since the data split of the triplet group ensures that no triplet in the form of $(h_0, r_0, t)$ exists in the training set, the conditional probability is not available.

In the experiments, we set the size of the embedding vectors of all entities and relations to be 20. We used the L1 distance in training. We set the other parameters as follows for all models: $K=1000$, $\gamma=1$, $\lambda=10$, $\alpha_r=\{1,1,1,1,1,10\}$, $\beta_r=\{15,10,10,15,10,0\}$, $\varepsilon_p=10^{-4}$, and $\varepsilon_n=10^{-13}$. The values of each relation of $\alpha_r$ and $\beta_r$ are in the sequence of *disease_to_medicine*, *disease_to_symptom*, *disease_to_operation*, *disease_to_laboratory*, *disease_to_examination*, and *upper_disease_to_lower_disease*.

### Evaluation Results

Figures 3, 4, and 5 compare the performance of TransX algorithms and the corresponding PrTransX algorithms, in terms of Hits@10, mean rank, and NDCG@10. The score for each algorithm is the average value on all of the triplets in the evaluation set. In our study, PrTransX performed better than its corresponding TransX model for X=E/H/R/D/Sparse in nearly all performance indicators, achieving a higher Hits@10 and NDCG@10 and lower mean rank.

**Figure 3.** Proportion of corrected entities ranked in the top 10 of the tested algorithms.

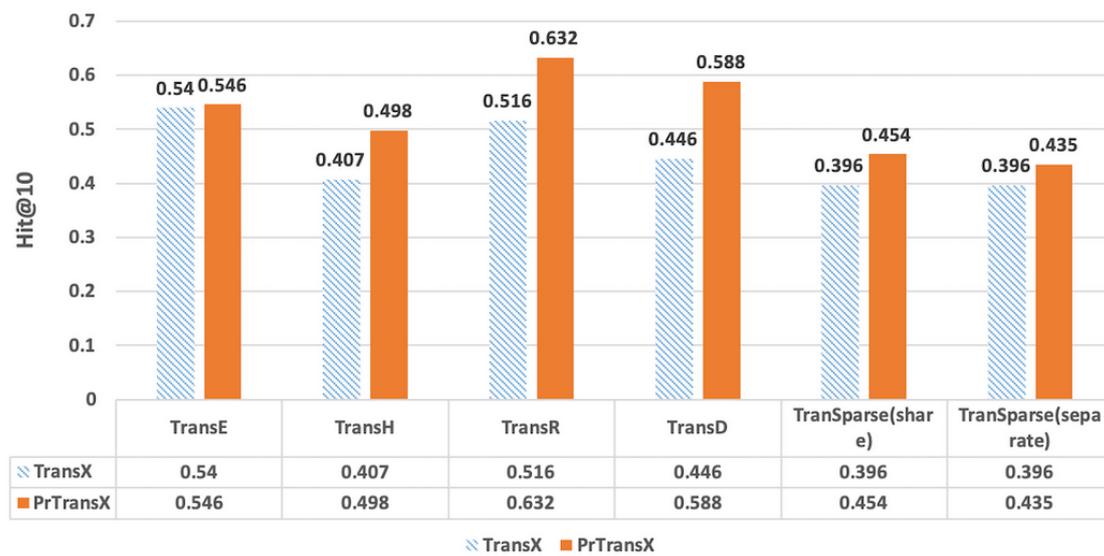

**Figure 4.** Mean rank of the tested algorithms.

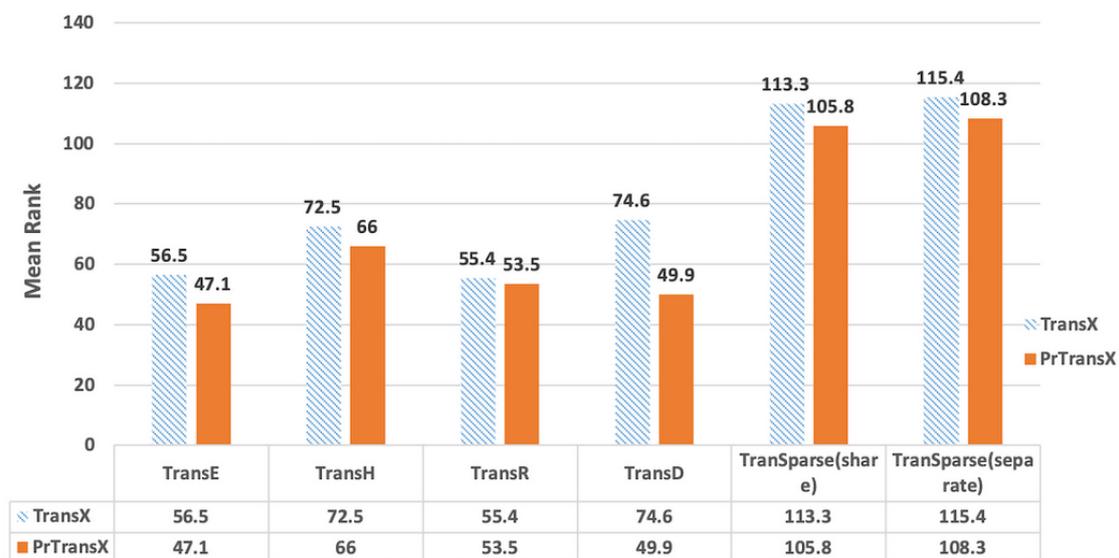

**Figure 5.** Normalized discounted cumulative gain of the top 10 predicted tail entities of the tested algorithms.

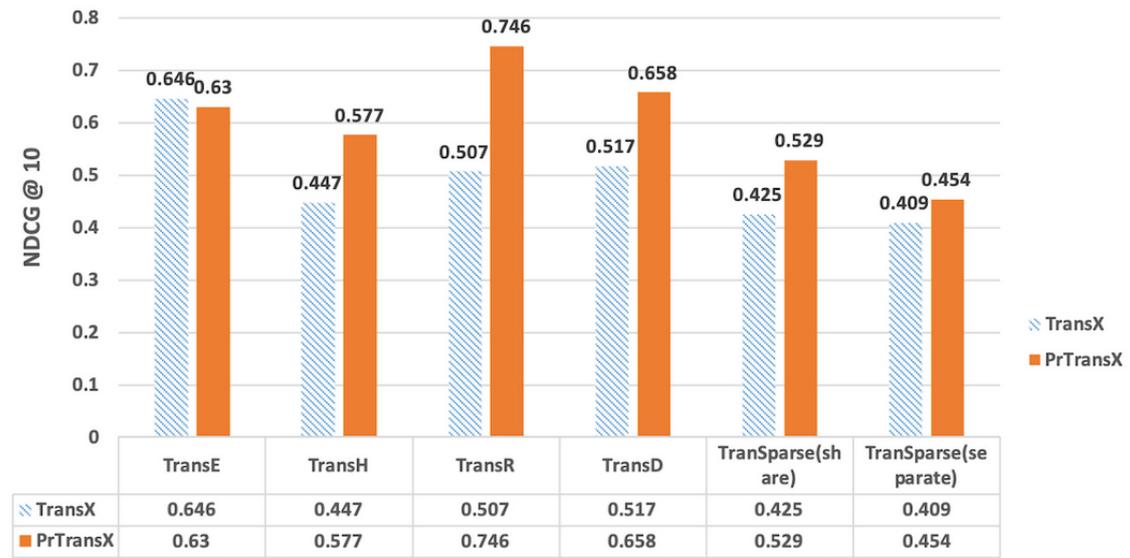

In addition to comparing the average score on the triplets for all the relations, we also examined the detailed performances for each relation.

In order to describe the distribution characteristics of entities by each relation in a quantitative manner, we calculated the *number of head entities per tail entity* by relations. For a given relation and tail entity, the *number of head entities per tail entity* is the number of distinct head entities that relate to the tail entity. Next, the numbers of head entities for each of the tail entities for a given relation form *the distribution of head entities numbers* for the relation.

Figure 6 illustrates the distribution of head entities numbers by different relations. For each box, the upper bound, the median line with a notch, and the lower bound represent the value at the 75%, 50%, and 25% percentiles of the data, respectively. The median number of relation *disease_to_examination* is near 150, and the 75% percentile value is above 350, which means that 50% of examination entities are related to near 150 diseases and 25% are related to more than 350 diseases. The average numbers of related diseases for other entities (ie, laboratory, symptom, medicine, and operation) are all less than the examination entity. The one with the fewest is operation.

**Figure 6.** Distribution of head entities numbers by different relations.

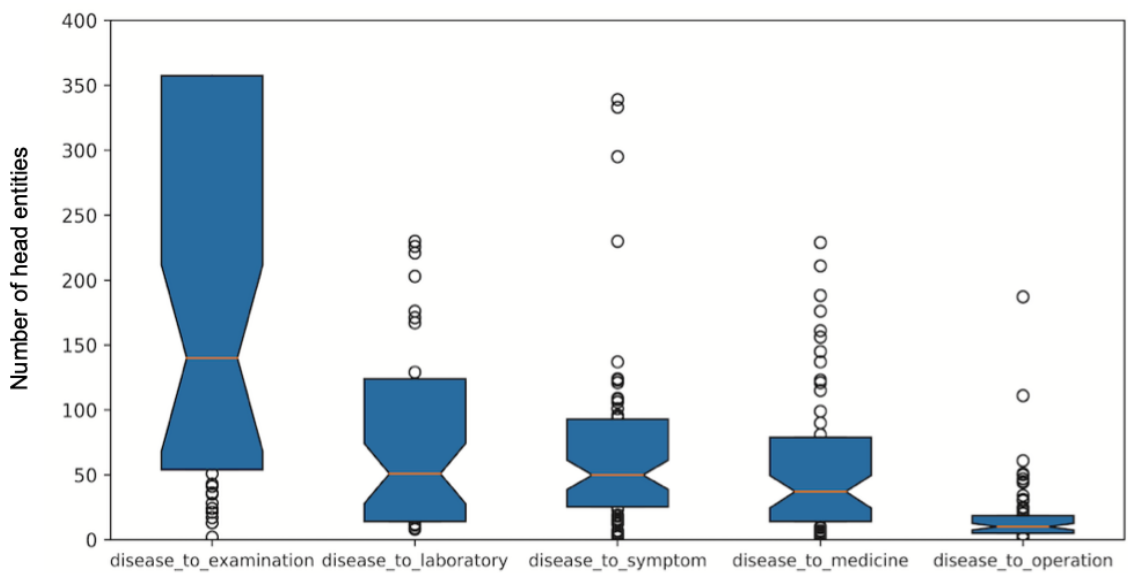

Figure 7 compares NDCG@10 of TransX with its corresponding PrTransX by different relations. Each subgraph represents the NDCG@10 comparison of TransX and PrTransX for a given X model on each of the 5 relations. The performance of PrTransX is drawn as a solid line, while the performance of TransX is drawn as a dashed line.

**Figure 7.** Comparison of normalized discounted cumulative gain of the top 10 predicted tail entities of TransX and PrTransX by different relations.

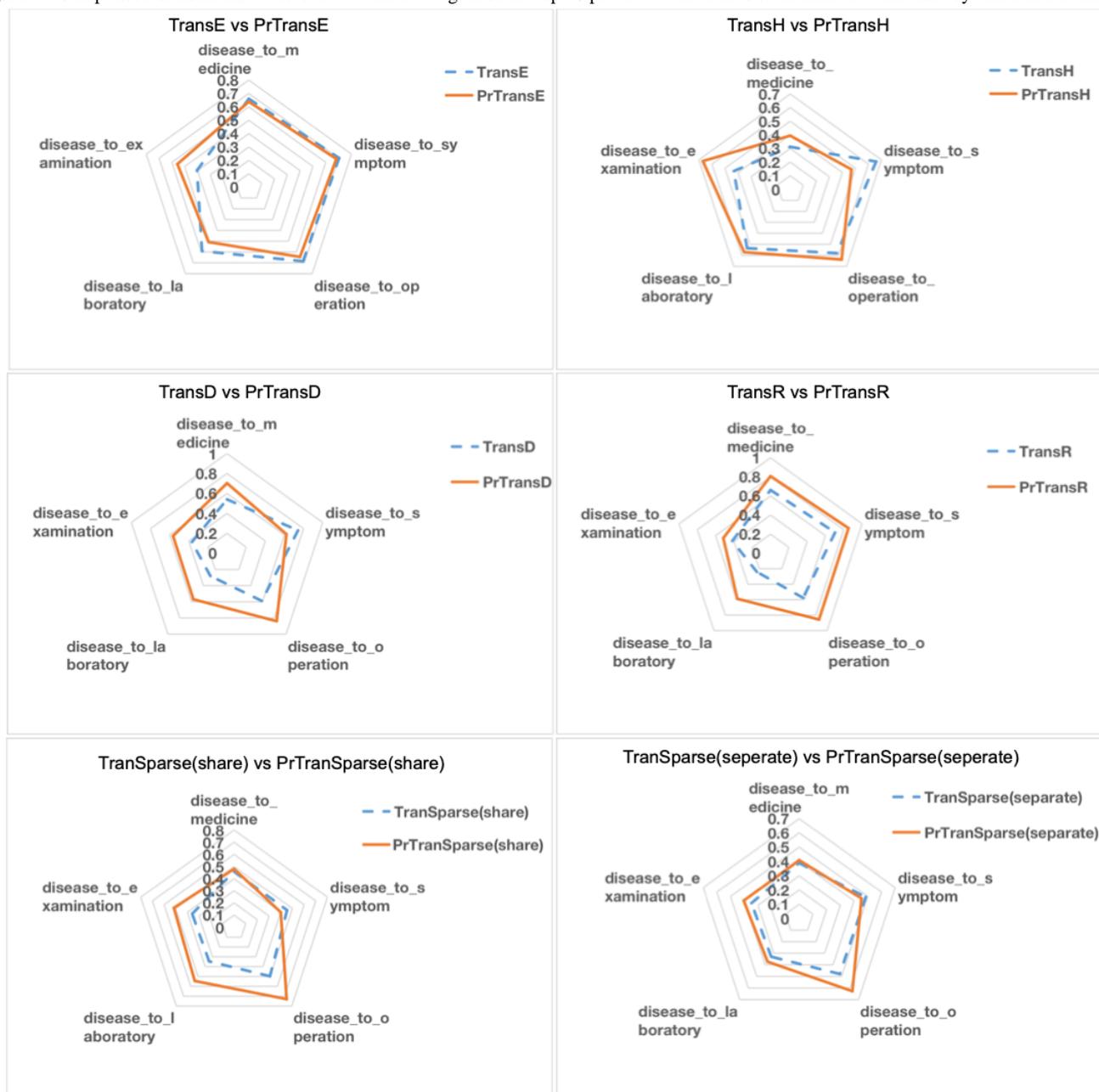

## Discussion

### Principal Findings

As Figures 3, 4, and 5 show, PrTransX performed better than the TransX model for X=E/H/R/D/Sparse in nearly all performance indicators, achieving a higher Hits@10 and NDCG@10 and lower mean rank. All these data proved that by adding probability-based loss into the translation-based loss functions, the obtained embedding vectors improved the link prediction performance. Furthermore, PrTransR performed the best across all PrTransX algorithms under Hits@10 and NDCG@10, and PrTransE performed better under mean rank.

The performance of TransE was generally better than that of TransH/R/D. This was due to the property of our medical KG. In the entire test set, there were only 5 different relations. The number of relations was so small that it was possible to train the embedding of all entities and relations in the same space to satisfy the training objective, which was similar to the result that the link prediction performance of TransH was worse than TransE on the WN18 data set used by Wang et al [6].

The results of TransE and PrTransE were quite similar under the Hits@10 and NDCG@10. In particular, the NDCG@10 of TransE was slightly better than that of PrTransE. Such similar performances were because TransE embedded all entities and relations into the same space. The probability-based training targets of the different relations were impossible to satisfy in the same space. The results of PrTransH/R/D show that the probability-based training targets introduced significant improvements in the Hits@10 and NDCG@10 over those measures of the TransH/R/D model.

These statistics in Figure 6 indicate that the relation *disease_to_examination* was not specific, since each examination entity was related to huge amount of disease entities on average. In contrast, *disease_to_operation* was the most specific relation, as the median number was 10. This conforms the medical common sense that the same examination is applicable to many types of diseases, but the same operation is applicable to only a few types of diseases.

Regarding the obvious performance difference for different relations shown in Figure 7, we can conclude that if the tail entity type is not specific (eg, examination), the link prediction on the evaluation data set is relatively difficult, since it is hard to train tail entities to fit distance requirements of a large amount of distinct head entities. Otherwise, if the tail entity type is specific to disease (eg, operation), the link prediction task would be relatively easier because the training objective is easier to be satisfied.

Overall, the polygons with solid lines (PrTransX) are larger than those of dashed lines (TransX), which means that PrTransX performed better. First, the best NDCG@10 score was achieved by PrTransR in the relation *disease_to_operation*. Moreover, in relation to *disease_to_operation*, PrTransD/Sparse(share)/Sparse(separate) also achieved significant improvements over the respective TransX. Such results echo the previous argument that the link prediction task would be relatively easier for specific relations than for unspecific relations. Second, even if the relation *disease_to_examination* is unspecific (difficult to predict), PrTransX outperformed TransX in this relation for all models.

## Limitations

This study had several limitations. First, the algorithm PrTransX can only obtain performance improvement on a probabilistic KG. For the KGs in general domains that have no probability over the relations, PrTransX algorithms are the same as the classical TransX algorithms. Second, the evaluation data set used by the link prediction task contained only 335 triplets from 25 triplet groups, since the labeling is quite labor-intensive.

## Future Studies

It should be noted that the application scenarios of knowledge embedding are far beyond link prediction. In state-of-the-art natural language processing research, such as ERNIE [10], significant improvements in various knowledge-driven tasks are achieved by using both text contexts and KGs to train word-embedding vectors. In the medical field, Zhao et al [11] also reported that embedding vectors from TransE and latent factor models could be used in a conditional random field to infer possible diagnoses based on laboratory test results and symptoms. Applying embedding vectors learned by PrTransX from a probabilistic medical KG into the clinical decision support system is worth exploring in the future.

## Conclusion

We proposed PrTransX to learn the embedding vectors of a probabilistic KG. We performed the study on a medical KG constructed from large-scale EMR data, and evaluation on link prediction indicated that the embedding learned by the PrTransX significantly outperformed that learned by corresponding TransX algorithms. We can conclude that the proposed PrTransX successfully incorporated the uncertainty of the knowledge triplets into the embedding vectors.

### Conflicts of Interest
None declared.

## Abbreviations

**EMR:** electronic medical record
**HITS@10:** proportion of corrected entities ranked in the top 10
**ICD-10:** International Classification of Diseases, Tenth Revision
**KG:** knowledge graph
**NDCG@10:** normalized discounted cumulative gain of the top 10 predicted tail entities
**PBE:** probabilistic belief embedding